\documentclass{article}

\usepackage[a4paper, total={6.7in, 9.7in}]{geometry}
\usepackage{cite}
\usepackage{url}
\usepackage{array}
\usepackage{amsmath,amssymb,amsfonts}
\usepackage{algorithm,algorithmic}
\usepackage{graphicx}
\graphicspath{{./figures/}}
\usepackage{textcomp}
\usepackage{xcolor}
\usepackage{booktabs}
\usepackage{multirow}
\usepackage{float}
\usepackage{longtable}
\usepackage{caption}
\usepackage{authblk}
\usepackage{tabularx}

\date{}

\title{\textit{PsyBridge}: A Hybrid Intelligent Framework for Multi-Dimensional Mental Health Assessment and Decision Support}

\author[1]{Sunil Wanjari\thanks{Corresponding author}}
\author[2]{Manish Thakre}
\author[1]{Aayushi Jayant Asole}
\author[1]{Sharwari Raut}
\author[3]{Kwabena Adu-Duodu}
\author[3]{Yinhao Li}
\author[1]{Stanly Wilson}

\affil[1]{St. Vincent Pallotti College of Engineering and Technology, Nagpur, India}
\affil[2]{Government Medical College (GMC), Chandrapur, India}
\affil[3]{Newcastle University, UK}

\begin{document}
\maketitle

\begin{abstract}
Mental health assessment commonly relies on isolated screening instruments or data-driven models that often lack interpretability and multi-dimensional integration. Existing approaches frequently focus on individual indicators such as depression or anxiety while providing limited support for comprehensive and explainable decision-making. To address this limitation, this study proposes \textit{PsyBridge}, a hybrid intelligent decision-support framework designed for multi-dimensional mental health assessment through the integration of clinically validated screening tools, cognitive evaluation, and personality profiling within a unified architecture. The proposed framework incorporates PHQ-9 and GAD-7 assessments alongside cognitive and behavioural indicators using a modular design and a weighted aggregation mechanism to generate interpretable mental health risk classifications and recommendations. To evaluate the framework, a semi-synthetic dataset consisting of 500 patient profiles representing varying severity levels was constructed based on clinically grounded score distributions. Experimental results demonstrate that \textit{PsyBridge} achieves an overall accuracy of 0.84, outperforming standalone PHQ-9 and GAD-7 assessments while improving precision, recall, and F1-score. Sensitivity analysis and ablation studies further indicate that integrating cognitive and personality components contributes to more stable classification performance and reduces inconsistencies in moderate-risk prediction. The findings suggest that \textit{PsyBridge} provides a scalable and interpretable approach for AI-assisted mental health decision support, particularly within digital healthcare and telehealth environments. The modular architecture facilitates future integration with adaptive machine learning techniques and real-world clinical systems. Future work will focus on validation using clinical datasets and development of data-driven weighting strategies to enhance generalisability and practical deployment.

\end{abstract}


\section{Introduction}

Mental health conditions, such as depression, anxiety, and cognitive impairments, represent a major and growing global health concern, affecting populations across diverse social and demographic contexts. According to reports by the World Health Organization, depression remains a leading cause of disability worldwide, underscoring the need for reliable and scalable assessment strategies\cite{whoWorldMental}.

Widely used clinical screening tools, including the Patient Health Questionnaire (PHQ-9) and the Generalised Anxiety Disorder scale (GAD-7), provide validated measures for identifying depression and anxiety \cite{kroenke_2001,spitzer_2006}. However, in many practical settings, these instruments are applied independently, resulting in fragmented evaluation processes. Such an approach often fails to incorporate additional influencing factors, including cognitive performance and personality traits, which can significantly affect mental health outcomes. Consequently, reliance on single-domain assessments may produce incomplete or inconsistent evaluations, particularly in high-demand or resource-constrained environments.

Advances in artificial intelligence have enabled the development of data-driven methods and decision support systems for mental health analysis \cite{shatte_2019,abd_2019}. Although machine learning techniques have shown promising predictive performance, their dependence on large datasets and limited interpretability can hinder their adoption in clinical practice \cite{torous_2021}. In contrast, rule-based and expert systems provide greater transparency but often lack the capability to integrate multiple assessment dimensions effectively\cite{samek_2019}.

These limitations highlight the need for hybrid and interpretable frameworks capable of combining heterogeneous information sources within a unified decision-making structure\cite{holzinger_2019}. Integrating clinical screening with cognitive assessment and personality profiling offers a more comprehensive representation of an individual’s mental health condition while preserving its interpretability. To address these limitations, this study introduces \textit{PsyBridge}, a hybrid intelligent framework for multidimensional mental health assessment and decision-making support. The proposed system combines validated clinical instruments with cognitive and personality-based inputs through a modular architecture supported by a weighted decision mechanism, thereby enabling consistent, interpretable, and scalable outcomes.

{The primary contributions of this work are as follows:}

\begin{itemize}

\item A novel multi-dimensional mental health assessment framework, \textit{PsyBridge}, is proposed to combine clinical screening, cognitive assessment, and personality profiling within a unified decision-support environment.

\item A weighted aggregation mechanism is introduced to integrate heterogeneous assessment outcomes into an interpretable mental health risk score while maintaining transparency and clinical relevance.

\item A structured risk classification and recommendation process is developed to transform assessment outcomes into actionable decision-support insights for mental health evaluation.

\item Extensive experimental validation demonstrates that the proposed framework consistently outperforms conventional PHQ-9-only and GAD-7-only screening approaches, highlighting the benefits of multi-dimensional assessment for mental health risk prediction.
\end{itemize}

{The remainder of this paper is organized as follows. Section~\ref{sec:literature} reviews the relevant literature on mental health assessment, clinical screening instruments, cognitive evaluation, personality profiling, and AI-enabled decision-support systems. Section~\ref{sec:methodology} presents the proposed \textit{PsyBridge} framework, including its system architecture, feature representation, weighted decision model, and algorithmic workflow. Section~\ref{sec:setup} describes the experimental setup, including dataset construction, baseline methods, evaluation metrics, and validation procedures. Section~\ref{sec:results} presents the experimental results and performance analysis through quantitative evaluation, confusion matrix analysis, graphical interpretation, ablation studies, sensitivity analysis, and module contribution assessment. Section~\ref{sec:limitations} discusses the implications of the findings, practical considerations, and limitations of the proposed framework. Finally, Section~\ref{sec:colclusion} concludes the paper and outlines directions for future research.}

\section{Literature Review} \label{sec:literature}

Research on digital mental health assessment has expanded considerably with advances in artificial intelligence (AI), machine learning, clinical decision support systems (CDSS), and digital healthcare technologies. Existing studies generally focus on four major domains: AI-based mental health prediction, explainable and interpretable AI, clinical decision-support frameworks, and large language model (LLM)-driven mental healthcare.

Traditional mental health assessment primarily depends on clinically validated screening instruments, such as the Patient Health Questionnaire (PHQ-9) and the Generalised Anxiety Disorder Scale (GAD-7), which are extensively used for evaluating depression and anxiety severity due to their reliability and clinical interpretability \cite{kroenke_2001,spitzer_2006}. Although these tools provide standardised assessments, they evaluate psychological conditions independently and do not capture broader cognitive or behavioural contexts.

Machine learning approaches have increasingly been applied to mental health analysis using questionnaire responses, behavioural patterns, social media data, and multimodal healthcare information. Previous studies have reported promising predictive performance for detecting depression, anxiety, and psychological disorders \cite{shatte_2019,chancellor_2020}. Recent systematic reviews further emphasise the expansion of AI applications in anxiety management and mental healthcare interventions \cite{das_2024,wajid_2025}. Despite strong predictive capabilities, many AI models operate as black-box systems that require large labelled datasets, thereby limiting transparency and clinical trust \cite{lee_2021}.

To address interpretability challenges, explainable artificial intelligence (XAI) approaches have been introduced to improve transparency in AI-driven decisions. Techniques such as SHAP and LIME provide post hoc explanations for predictive models \cite{Lundberg_2017,ribeiro_2016}, whereas healthcare-oriented explainability frameworks emphasise causability and clinical interpretability \cite{holzinger_2019,samek_2019}. However, most explainability mechanisms remain supplementary rather than being inherently integrated into decision-support architectures.

Clinical decision-support systems represent another major research direction. Recent studies have explored AI-assisted psychiatric support, mental disorder prediction, and digital intervention systems \cite{tutun_2023,bertl_2022}. Systematic reviews indicate the increasing adoption of AI-enabled decision-support systems within psychiatric and mental healthcare services \cite{golden_2024,gu_2024}. Nevertheless, implementation barriers persist, including insufficient interpretability, privacy concerns, and limited integration with clinical workflows \cite{auf_2025,poon_2025}.

Digital psychiatry platforms and AI-enabled mental healthcare applications have improved accessibility through mobile applications, chatbots, and remote monitoring technologies \cite{torous_2021,abd_2019}. More recently, large language models (LLMs) have emerged as promising tools for conversational assessment, psychological support, and scalable mental healthcare delivery \cite{lai_2023,hua_2025}. However, studies emphasise the limitations related to reliability, explainability, ethical concerns, and application boundaries \cite{yang_2023,yang_2025}.

Ethical deployment, privacy preservation, and shared decision-making remain critical challenges for AI-supported healthcare systems. Existing studies highlight the importance of trust, clinician--patient interaction, privacy protection, and human-centred decision support for successful implementation \cite{lorenzini_2023,lee_2022,slade_2017,makoul_2006}.




\begin{table*}[t]
\centering
\footnotesize
\renewcommand{\arraystretch}{1.1}

\caption{Comparative Analysis of Existing Mental Health Assessment Approaches and Positioning of \textit{PsyBridge}}
\label{tab:existing_assessment}

\begin{tabularx}{\textwidth}{
p{2.0cm}
p{2.3cm}
X
p{1.3cm}
p{1.6cm}
p{1.5cm}
X
}
\toprule

\textbf{Study} &
\textbf{Primary Approach} &
\textbf{Assessment Focus} &
\textbf{Explain-able} &
\textbf{Multi-Dimensional } &
\textbf{Clinical Decision Support} &
\textbf{Key Limitation} \\

\midrule

\cite{kroenke_2001} & PHQ-9 & Depression Screening & High & No & High & Limited to depression \\

\cite{spitzer_2006} & GAD-7 & Anxiety Screening & High & No & High & Limited to anxiety \\

\cite{shatte_2019} & Machine Learning & Mental health prediction & Low & Partial & Limited & Black-box models \\

\cite{torous_2021} & Digital Psychiatry & Apps and monitoring & Medium & Partial & Moderate & Limited explainability \\

\cite{holzinger_2019}  & Explainable AI & Clinical AI interpretation & High & Partial & Moderate & Limited integration \\

\cite{ribeiro_2016} & LIME & Model explanation & High & No & Limited & Post hoc explanation \\

\cite{Lundberg_2017} & SHAP & Explainable prediction & High & No & Limited & Not clinically integrated \\

\cite{tutun_2023} & AI Decision Support & Mental disorder prediction & Medium & Partial & Moderate & Limited behavioural context \\

\cite{bertl_2022} & Psychiatric CDSS & Psychiatry systems & Medium & Partial & Moderate & Implementation barriers \\

\cite{golden_2024} & AI Clinical Support & Mental healthcare & Medium & Partial & High & Limited cognitive integration \\

\cite{gu_2024} & Mental Health CDSS & Helpline services & Medium & Partial & High & Limited personalization \\

\cite{lee_2021} & AI in Mental Healthcare & Clinical AI applications & Medium & Partial & Moderate & Adoption challenges \\

\cite{lai_2023} & LLM-based Support & Psychological assistance & Low--Medium & Partial & Limited & Reliability issues \\

\cite{yang_2025} & LLM Review & Mental healthcare & Medium & Partial & Limited & Boundary limitations \\

\cite{wajid_2025} & Systematic Review & AI in mental health & Medium
& Partial & Moderate & Lack of unified framework \\

{\textit{PsyBridge} (Proposed)}
& {Hybrid Intelligent Decision Framework}
& {Clinical + Cognitive + Personality Assessment}
& {High}
& {Yes}
& {High}
& {Addresses integration gap} \\

\bottomrule
\end{tabularx}
\end{table*}

{In addition to the comparative analysis presented in Table~\ref{tab:existing_assessment}, it is important to critically examine the limitations of existing approaches in greater depth. Machine learning-based systems, while demonstrating strong predictive performance, often rely heavily on large-scale labelled datasets and complex feature engineering processes. This dependency limits their applicability in real-world clinical settings, in which labelled data may be scarce or heterogeneous. Furthermore, many existing studies focus primarily on single-domain analyses, such as depression or anxiety detection, without considering the broader psychological and behavioural context. This reductionist approach may lead to incomplete or biased assessments, particularly in cases where multiple factors interact simultaneously. Recent research has emphasized the importance of multi-modal and multi-dimensional analysis in mental health, highlighting the need for systems that can integrate diverse data sources.}

{Explainable artificial intelligence (XAI) techniques have been introduced to address the transparency limitations of black-box models; however, these methods often provide post hoc explanations rather than inherently interpretable decision processes. As a result, there remains a gap between model performance and clinical usability. In contrast, the proposed \textit{PsyBridge} framework adopts an inherently interpretable design, combining clinically validated screening tools with cognitive and personality-based assessments. This approach enhances predictive capability and aligns with the requirements of clinical decision-making, where interpretability and reliability are critical. As shown in Table~\ref{tab:existing_assessment}, existing approaches either focus on performance without interpretability or provide explainability without comprehensive integration. In particular, most systems lack the ability to combine clinical screening, cognitive assessment, and personality profiling within a single framework. To address these limitations, the proposed \textit{PsyBridge} framework adopts a hybrid intelligent approach that integrates multiple assessment modalities into a unified and interpretable decision-support system.}

The \textit{PsyBridge} framework integrates clinically validated psychological screening, cognitive assessment, personality profiling, explainable risk computation, and unified recommendation support, within a single decision-support architecture. This multi-dimensional approach enables a more comprehensive, interpretable, and clinically meaningful assessment of mental health risk.


\section{Methodology and System Architecture} \label{sec:methodology}

The \textit{PsyBridge} framework is designed as a hybrid intelligent system that integrates multiple dimensions of mental health assessment into a unified decision-support pipeline.

\subsection{Methodological Workflow}
\begin{figure}[H]
\centering
\includegraphics[width=\linewidth]{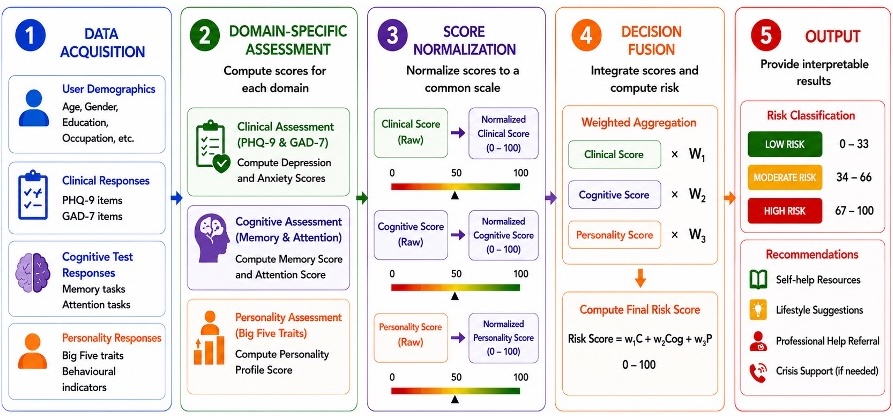}
\caption{Workflow of \textit{PsyBridge}}
\label{fig:workflow}
\end{figure}

The methodological workflow of \textit{PsyBridge}, as illustrated in Figure~\ref{fig:workflow}, represents a structured sequence of operations that transform raw user input into interpretable mental health risk assessments. The process begins with data acquisition through user responses, followed by score computation across clinical, cognitive, and personality modules. These scores are then normalised to a common scale to ensure consistency across heterogeneous inputs. Subsequently, a weighted aggregation mechanism is applied, in which each component contributes proportionally to the final risk score based on its relative importance. This step enables the integration of multiple assessment dimensions into a unified decision framework. The aggregated score is then mapped to predefined risk categories, facilitating clear interpretation and actionable insights. The final output includes both risk level classification and recommendations, supporting practical decision-making in clinical and digital health contexts.

{The decision process within the \textit{PsyBridge} framework is organized into five sequential stages. The first stage, data acquisition, involves collecting user responses from clinical screening instruments, cognitive assessments, and personality profiling questionnaires. In the second stage, domain-specific assessment, individual scores are computed for the clinical, cognitive, and personality domains using their respective evaluation mechanisms. The third stage performs score normalisation to transform the heterogeneous assessment outputs into a common scale, ensuring consistency and comparability across domains. The fourth stage is the decision fusion stage which integrates the normalised scores through a weighted aggregation mechanism to compute the overall mental health risk score and support subsequent risk classification and recommendation generation. Finally the results are generated.}

\subsection{System Architecture}

\begin{figure}[H]
\centering
\includegraphics[width=\linewidth]{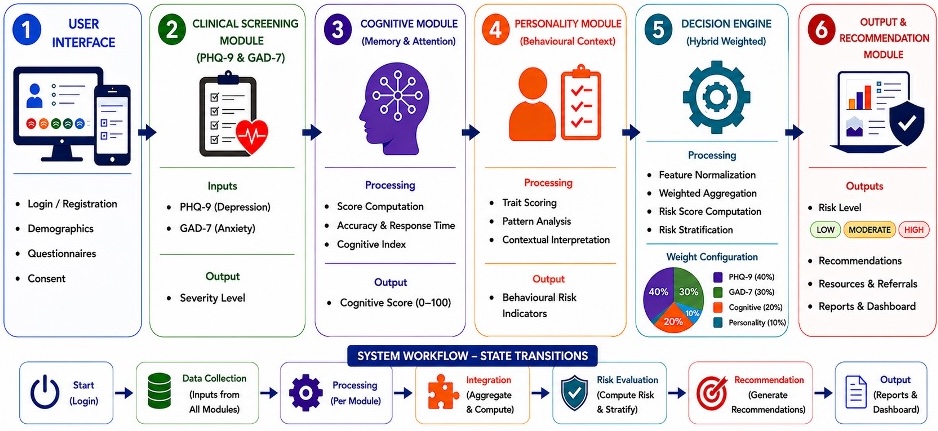}
\caption{System Architecture of \textit{PsyBridge}}
\label{fig:architecture}
\end{figure}

{The overall architecture of \textit{PsyBridge} is illustrated in Figure~\ref{fig:architecture}. The system follows a modular design integrating clinical, cognitive, and personality components within a unified decision-support framework. The proposed \textit{PsyBridge} framework follows a modular architecture designed to facilitate comprehensive mental health assessment through the integration of multiple psychological and cognitive dimensions. The system comprises five primary modules: the User Interface, which captures structured patient responses; the Clinical Screening Module, which uses symptom-specific instruments such as PHQ-9 and GAD-7 measures; the Cognitive Module, which evaluates memory and attentional-related cognitive functions; the Personality Module, which provides behavioural and emotional insights through personality profiling; and the Decision Engine, which integrates information from all modules using a hybrid weighted aggregation mechanism to generate an overall risk assessment and personalised recommendations. This modular design enhances system scalability, interpretability, and maintainability while enabling seamless integration of future learning and predictive analytics components. Furthermore, the architecture supports deployment across clinical, academic, and digital health environments, making it suitable for both research and real-world mental health decision-support applications.}

\subsection{Module-wise Processing}
{The \textit{PsyBridge} framework performs mental health assessment through multiple interconnected domain-specific modules, where each module contributes complementary information to the overall decision-making process. The clinical screening module evaluates depression and anxiety severity using standardised assessment instruments such as PHQ-9 and GAD-7\cite{kroenke_2001, spitzer_2006}. The cognitive assessment module captures memory retention and attentional consistency indicators that are often associated with psychological stress and cognitive impairment. In parallel, the personality profiling module provides supplementary behavioural context through MBTI-based personality mapping, enabling contextual interpretation of user responses. The outputs generated by these modules are processed by the decision engine, which performs weighted multi-factor fusion to combine heterogeneous psychological indicators into a unified and interpretable risk score. Unlike conventional isolated screening approaches, \textit{PsyBridge} integrates behavioural, cognitive, and clinical dimensions within a single hybrid assessment framework. As shown in Table~\ref{tab:function_rule}, this integrated processing enhances assessment consistency, contextual understanding, and interpretability of the final decision output.}

\begin{table}[H]
\centering
\caption{Functional Role of Each Module}
\label{tab:function_rule}
\begin{tabular}{lll}
\toprule
Module & Function & Output \\
\midrule
Clinical Screening & PHQ-9, GAD-7 scoring & Depression \& Anxiety scores \\
Cognitive Assessment & Digit Span Test & Cognitive score \\
Personality Profiling & MBTI mapping & Personality influence \\
Decision Engine & Weighted aggregation & Final risk score \\
\bottomrule
\end{tabular}
\end{table}

\subsection{Score Normalization}

To ensure consistency and comparability across heterogeneous input features, all module outputs are normalised to a common scale in the range $[0,1]$. This step is essential because the input variables, such as PHQ-9 (0–27), GAD-7 (0–21), cognitive scores, and personality indicators, originate from different measurement scales and cannot be directly aggregated. The normalisation process is performed using min–max scaling\cite{han_2011} defined as:

\begin{equation}
x' = \frac{x - x_{\min}}{x_{\max} - x_{\min}}
\end{equation}

where $x$ represents the original score, $x_{\min}$ and $x_{\max}$ denote the minimum and maximum possible values of the respective feature, and $x'$ is the normalised value. This transformation preserves the relative differences between scores while ensuring that all features contribute proportionally to the weighted aggregation process. Without normalisation, features with larger numerical ranges could dominate the decision model, leading to biased risk estimation. By mapping all inputs to a unified scale, the normalisation step enhances the stability, fairness, and interpretability of the overall decision-making process.

\subsection{Weighted Decision Model}

The final mental health risk score in the \textit{PsyBridge} framework is computed using a weighted aggregation of normalised inputs obtained from clinical, cognitive, and personality assessment modules. This approach enables the integration of heterogeneous features into a single interpretable metric while preserving the relative importance of each component.
The risk score is formally defined as a linear combination of the normalised feature values as follows:

\begin{equation}
\text{Risk Score} =
w_1S_{\text{PHQ9}} +
w_2S_{\text{GAD7}} +
w_3S_{\text{Cog}} +
w_4S_{\text{Pers}}
\end{equation}

where $S_{\text{PHQ9}},\; S_{\text{GAD7}},\; S_{\text{Cog}},\; S_{\text{Pers}} \in [0,1]$ represent normalized scores, and $w_i$ are the corresponding weights such that

\begin{equation}
\sum_{i=1}^{4} w_i = 1
\end{equation}

The weights were assigned based on clinical relevance and empirical reasoning. The PHQ-9 and GAD-7 scores were given higher weights (0.4 and 0.3, respectively) because they are widely validated clinical instruments for assessing depression and anxiety severity\cite{kroenke_2001, spitzer_2006}. The cognitive component was assigned a moderate weight (0.2), reflecting its supportive role in identifying underlying impairments associated with mental health conditions. The personality component was assigned a lower weight (0.1) because it primarily provides contextual information rather than direct diagnostic evidence.

The use of a linear weighted model ensures computational efficiency and interpretability. Each component’s contribution to the final score can be explicitly traced, allowing both clinicians and users to understand how different factors influence the overall assessment. This transparency is particularly important in healthcare applications, where explainability is essential for trust and adoption. Furthermore, the weighted aggregation framework provides flexibility for future extensions. The weights can be adjusted or learned dynamically using data-driven approaches, enabling the model to adapt to different populations or clinical settings while maintaining its interpretable structure.



{The normalized feature values $S_{\text{PHQ9}}$, $S_{\text{GAD7}}$, $S_{\text{Cog}}$, and $S_{\text{Pers}}$ are obtained using the min--max normalization procedure described in the previous subsection. These normalised values represent depression severity, anxiety severity, cognitive assessment performance, and personality-based behavioural context, respectively. The weighted aggregation mechanism combines these heterogeneous indicators into a unified risk score while preserving the relative contribution of each assessment dimension. To facilitate interpretation and decision support, the computed risk score is mapped to predefined risk categories. Scores in the range $0.0 \leq \text{Risk Score} < 0.3$ are classified as \textbf{Low Risk}, indicating minimal evidence of mental health concerns. Scores in the range $0.3 \leq \text{Risk Score} < 0.6$ are classified as \textbf{Moderate Risk}, suggesting the presence of potential psychological distress that may require monitoring or supportive intervention. Scores greater than or equal to $0.6$ are classified as \textbf{High Risk}, indicating an elevated likelihood of significant mental health concerns and the potential need for professional assessment or intervention.}

\subsection{Weight Assignment Justification}
{The weighting configuration adopted in \textit{PsyBridge} reflects the relative clinical importance and contextual contribution of each assessment component within the mental health evaluation process. Rather than assigning equal importance to all inputs, the proposed framework prioritises clinically validated screening instruments while retaining cognitive and behavioural indicators as complementary sources of contextual information. As summarised in Table~\ref{tab:weight_justification}, greater emphasis is assigned to clinically validated screening instruments because of their established reliability and widespread use in mental healthcare, whereas supportive contextual indicators receive lower weights. Table~\ref{tab:weight_justification} demonstrates that the assigned weights are aligned with the clinical relevance and evidential strength of each assessment component, ensuring that the final risk score remains both interpretable and clinically grounded. The selected weighting scheme prioritises clinically grounded indicators while preserving interpretability and allowing incorporation of contextual behavioural information. Future work may investigate adaptive weighting strategies through clinician consensus or data-driven optimization methods.}

\begin{table}[h]
\centering
\caption{Clinical Justification for Weight Assignment}
\label{tab:weight_justification}
\begin{tabular}{|p{3cm}|c|p{7cm}|p{2.5cm}|}
\hline
\textbf{Assessment Component} &
\textbf{Weight} &
\textbf{Rationale} &
\textbf{Clinical   Importance} \\
\hline

PHQ-9 & 0.40 & Standard depression screening with strong empirical validation & High \\ \hline

GAD-7 & 0.30 & Widely used anxiety assessment tool & High \\ \hline

Cognitive Assessment & 0.20 & Provides supportive information on memory and attention & Moderate \\ \hline

Personality Profiling & 0.10 & Behavioural context modifier rather than diagnostic indicator & Supportive \\ \hline
\end{tabular}
\end{table}

\subsection{Risk Classification}
The weighted aggregation model combines multiple assessment dimensions into a unified risk score, enabling a balanced representation of clinical and behavioral factors. The assigned weights reflect the relative importance of each component in mental health evaluation. Clinical screening scores (PHQ-9 and GAD-7) are assigned higher weights (0.4 and 0.3, respectively) due to their strong empirical validation and widespread clinical use in assessing depression and anxiety severity\cite{kroenke_2001, spitzer_2006}. The cognitive component is assigned a moderate weight (0.2), as cognitive impairment is often associated with underlying mental health conditions but may not independently determine risk severity. The personality factor is assigned a lower weight (0.1), serving as a contextual modifier that influences vulnerability patterns rather than acting as a primary diagnostic indicator.

The aggregated risk score provides a continuous measure that captures the combined influence of all the assessment dimensions. Higher values indicate an increased likelihood of mental health risks, whereas lower values correspond to minimal or no risk. This continuous scoring approach enables flexible threshold-based classification and supports nuanced interpretations beyond binary decision-making. Furthermore, the weighted model ensures interpretability by explicitly revealing the contribution of each component to the final results. This transparency is particularly important in clinical and decision-support contexts, where understanding the reasoning behind a prediction is essential for trust and adoption of the model.
{To facilitate practical interpretation and intervention planning, the continuous risk score is mapped to predefined risk categories. As shown in Table~\ref{tab:risk_classification}, the proposed framework classifies individuals into low-, moderate-, and high-risk groups based on score thresholds derived from the normalized risk scale.}
\begin{table}[H]
\centering
\caption{Risk Classification Levels}
\label{tab:risk_classification}
\begin{tabular}{ll}
\toprule
Risk Level & Score Range \\
\midrule
Low & 0.0 -- 0.3 \\
Moderate & 0.3 -- 0.6 \\
High & 0.6 -- 1.0 \\
\bottomrule
\end{tabular}
\end{table}

\subsection{Interpretability and Complexity}
Each component contributes explicitly to the final risk score, enabling both users and clinicians to understand how individual factors influence the overall outcome\cite{samek_2019}. The interpretability of the model further supports effective error analysis and system validation. By examining the contribution of each module, inconsistencies or unexpected outcomes can be systematically identified and addressed, thereby improving reliability and robustness and transparent and traceable reasoning\cite{samek_2019, holzinger_2019}.

From a computational perspective, the framework exhibits linear complexity, O(n), where n represents the number of input features. Because the model relies on simple operations, such as normalisation and weighted aggregation, it requires minimal computational resources and supports real-time execution. Furthermore, the modular design enhances scalability and extensibility, allowing seamless integration into mobile health systems, telepsychiatry platforms, and resource-constrained clinical environments. Overall, the combination of interpretability and computational efficiency positions \textit{PsyBridge} as a practical and reliable decision-support system for real-world mental health assessment.

\section{Experimental Setup} \label{sec:setup}
An evaluation of the proposed \textit{PsyBridge} framework approach was conducted to determine its effectiveness, a structured experimental design was developed, focusing on performance comparison with traditional screening approaches and validation of the integrated decision-support mechanism.

\subsection{Dataset Construction}
{Due to the absence of publicly available datasets that simultaneously include clinical screening scores, cognitive assessment measures, and personality profiling information, a semi-synthetic dataset was constructed for evaluation purposes. The dataset generation process was grounded in clinically validated scoring distributions reported in PHQ-9 and GAD-7 literature to ensure realistic representation of mental health conditions. A total of 500 patient profiles were generated. Each profile consisted of depression severity scores, anxiety severity scores, cognitive assessment outcomes, and personality-based behavioural indicators. The generated dataset maintained a balanced distribution across low-, moderate-, and high-risk categories to facilitate robust evaluation of the proposed framework. To enhance realism, controlled variability was introduced into the cognitive and personality components while preserving consistency with clinically established screening patterns. This design enables the evaluation to reflect realistic decision-making scenarios while maintaining alignment with established clinical scoring frameworks.}

\subsection{Dataset Characteristics and Limitations}
The constructed dataset reflects clinically plausible distributions based on established scoring guidelines; however, it is important to acknowledge certain limitations. The use of a semi-synthetic dataset, while enabling controlled experimentation, may not fully capture the variability and complexity of real-world patient data. Factors such as reporting bias, environmental influences, and comorbid conditions are difficult to model accurately in a simulated environment.

To mitigate these limitations, care was taken to ensure that the generated data aligned with clinically validated score ranges and realistic variability patterns. Additionally, the balanced distribution across risk categories helps prevent bias toward any particular class, ensuring a fair evaluation of the proposed model. Future work will focus on validating the framework using real-world clinical datasets and longitudinal data, which would enable a more comprehensive assessment of system performance and generalisability.

\subsection{Feature Representation}
Each patient profile is represented using multiple psychological and behavioural features extracted from the assessment modules. Since the features originate from heterogeneous scales, all values are normalised to a common range of 0–1 prior to aggregation to ensure numerical consistency during risk computation. {Together, these features capture complementary clinical, cognitive, and behavioural dimensions of mental health, enabling a more comprehensive assessment than traditional single-domain screening approaches. Table~\ref{tab:feature} summarises the feature representation used within the \textit{PsyBridge} framework. The selected feature set provides a balanced representation of symptom severity, cognitive functioning, and behavioural context, forming the basis for the weighted decision model described in the subsequent section.}

\begin{table}[H]
\centering
\caption{Feature Representation}
\label{tab:feature}
\begin{tabular}{ll}
\toprule
Feature & Description \\
\midrule
PHQ-9 Score & Depression severity (0--27) \\
GAD-7 Score & Anxiety severity (0--21) \\
Cognitive Score & Digit Span normalized score \\
Personality Score & Encoded MBTI influence factor \\
\bottomrule
\end{tabular}
\end{table}

\subsection{Baseline Methods}
{To evaluate the effectiveness of the proposed framework, \textit{PsyBridge} was benchmarked against widely adopted standalone mental health screening approaches, including PHQ-9-based depression assessment and GAD-7-based anxiety assessment. These methods represent conventional clinical screening practices that rely on a single assessment dimension for risk evaluation. Comparing \textit{PsyBridge} with these baseline approaches highlights the potential benefits of integrating clinical screening, cognitive assessment, and personality profiling within a unified decision-support framework. This comparative analysis demonstrates the contribution of multi-dimensional assessment in improving the comprehensiveness and reliability of mental health risk evaluation.}

\subsection{Evaluation Metrics}
{The performance of the proposed {\textit{PsyBridge}} framework is evaluated using widely accepted classification metrics to assess its effectiveness in mental health risk prediction. Accuracy measures the overall correctness of the classification outcome by determining the proportion of correctly classified instances among all cases. Precision evaluates the proportion of correctly identified positive cases among all cases predicted as positive, reflecting the reliability of positive predictions. Recall measures the system's ability to identify actual positive cases, indicating its effectiveness in detecting individuals at risk. The F1-score, computed as the harmonic mean of precision and recall, provides a balanced measure of classification performance by simultaneously considering both false positives and false negatives. These metrics collectively offer a comprehensive evaluation of the framework's predictive capability and reliability.}


\subsection{Implementation Details}

The \textit{PsyBridge} framework was implemented using a modular architecture comprising clinical screening, cognitive assessment, personality profiling, score normalization, and decision fusion modules. Each module was designed to operate independently, enabling flexibility, scalability, and future integration with advanced machine learning components. The experimental setup was established in a controlled simulation environment using a semi-synthetic dataset containing 500 patient profiles. The dataset was generated from clinically validated scoring distributions derived from PHQ-9 and GAD-7 literature and was balanced across low-, moderate-, and high-risk categories. Controlled variability was introduced to emulate realistic patient response patterns and assessment outcomes. The clinical module computed depression and anxiety severity scores, while the cognitive and personality modules generated corresponding assessment scores. To ensure comparability across domains, all scores were normalized to a common scale ranging from 0 to 1. A weighted decision fusion mechanism then combined the normalized outputs using predefined weights assigned to clinical, cognitive, and personality components to produce a unified mental health risk score.

The implementation was evaluated on a standard computing platform equipped with an Intel Core i5 processor, 8~GB RAM, and Python-based data processing libraries. Performance assessment was conducted using accuracy, precision, recall, and F1-score metrics. To demonstrate the effectiveness of the proposed multidimensional framework, \textit{PsyBridge} was benchmarked against conventional screening approaches based solely on PHQ-9 and GAD-7 assessments. The experimental setup was designed to ensure reproducibility, facilitate systematic evaluation of the decision fusion mechanism, and demonstrate the benefits of integrating clinical, cognitive, and personality assessments within a unified and explainable mental health decision-support framework.

\section{Result Analysis} \label{sec:results}
This section presents the performance evaluation of the proposed {\textit{PsyBridge}} framework using both quantitative metrics and graphical analysis. In addition to standard evaluation measures, a confusion matrix is included to provide deeper insight into classification performance.

\subsection{Quantitative Results} 

The performance of {\textit{PsyBridge}} was evaluated using standard classification metrics, including accuracy, precision, recall, and F1-score. Ground truth labels were assigned according to clinically established PHQ-9 and GAD-7 severity thresholds, where moderate-to-severe assessment outcomes were mapped to elevated mental health risk categories. The evaluation was conducted to compare the proposed multi-dimensional assessment framework with conventional single-dimensional screening approaches. The resulting risk categories were mapped to low-, moderate-, and high-risk classes to support multi-class classification and comparative evaluation.



\begin{figure}[H]
\centering
\includegraphics[width=\linewidth]{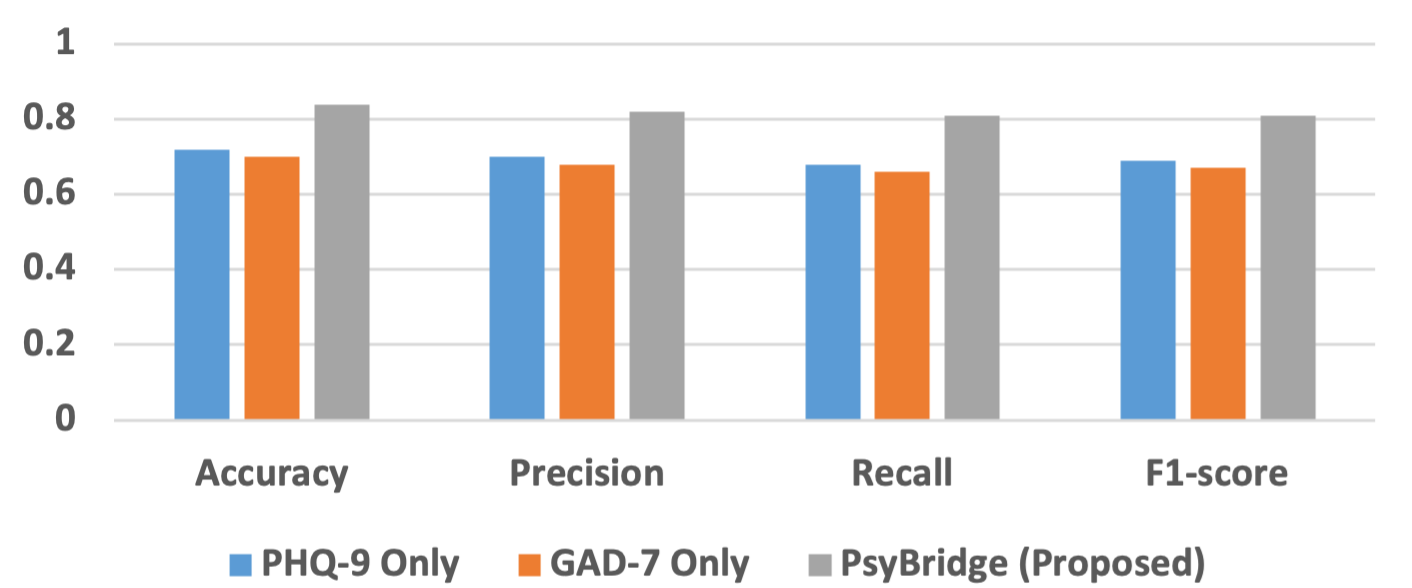}
\caption{Performance Comparison using Accuracy, Precision, Recall, and F1-score}
\label{fig:performance}
\end{figure}

{The results presented in Figure~\ref{fig:performance} demonstrate that the proposed {\textit{PsyBridge}} framework consistently outperforms the baseline single-dimensional screening approaches across all evaluation metrics. Specifically, {\textit{PsyBridge}} achieves an accuracy of 0.84 compared to 0.72 and 0.70 for the PHQ-9-only and GAD-7-only approaches, respectively. Similar improvements are observed for precision, recall, and F1-score, indicating enhanced predictive reliability and balanced classification performance. These improvements can be attributed to the integration of clinical screening, cognitive assessment, and personality profiling, which collectively provide a more comprehensive representation of an individual's mental health status than isolated symptom-based assessments. The findings support the central hypothesis of this study that multi-dimensional assessment improves the accuracy and robustness of mental health risk evaluation compared with traditional single-domain screening methods.}

\subsection{Confusion Matrix Analysis}
{To further evaluate classification performance, a confusion matrix was constructed for the proposed {\textit{PsyBridge}} framework, as shown in Table~\ref{tab:confusion}. The results demonstrate a strong concentration of correctly classified instances along the main diagonal, with 150 low-risk, 140 moderate-risk, and 160 high-risk cases correctly identified. Misclassifications are relatively infrequent and primarily occur between neighbouring risk categories. For example, 10 low-risk instances were classified as moderate risk, while 8 moderate-risk instances were classified as high risk. Such errors are expected in mental health assessment scenarios where symptom severity often exists on a continuum rather than within strictly separated categories. Importantly, very few low-risk cases were directly classified as high risk, and vice versa, indicating that the framework effectively preserves separation between distinct risk levels. These findings provide further evidence of the robustness and reliability of the proposed multi-dimensional assessment approach. The confusion matrix results complement the quantitative evaluation metrics by providing a detailed view of the distribution of classification errors across risk categories.}

\begin{table}[H]
\centering
\caption{Confusion Matrix for {\textit{PsyBridge}}}
\label{tab:confusion}
\begin{tabular}{lccc}
\toprule
Actual $\backslash$ Predicted & Low & Moderate & High \\
\midrule
Low & 150 & 10 & 5 \\
Moderate & 12 & 140 & 8 \\
High & 6 & 9 & 160 \\
\bottomrule
\end{tabular}
\end{table}

\subsection{Graphical Interpretation}

\begin{figure}[H]
\centering
\includegraphics[width=\linewidth]{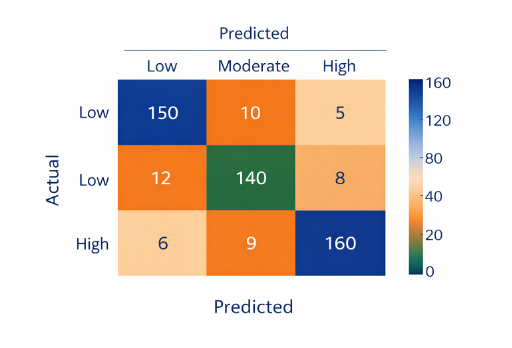}
\caption{Confusion Matrix Visualization}
\label{fig:confusion}
\end{figure}

{The confusion matrix provides a detailed view of classification outcomes by comparing predicted risk categories against the corresponding ground-truth labels. Figure~\ref{fig:confusion} illustrates the classification behaviour of the proposed {\textit{PsyBridge}} framework across the three risk categories. A strong concentration of instances along the diagonal indicates high classification accuracy, with 150 low-risk, 140 moderate-risk, and 160 high-risk cases correctly identified. Most misclassifications occur between neighbouring risk categories, such as low-to-moderate or moderate-to-high, which is expected given the gradual transition of symptom severity across mental health conditions. Importantly, very few low-risk cases are directly classified as high risk, and vice versa, demonstrating the framework's ability to preserve meaningful separation between distinct risk levels. These findings provide additional evidence that integrating clinical, cognitive, and personality-based indicators contributes to robust and reliable mental health risk assessment.}

\subsection{Ablation Study}

To evaluate the contribution of individual {\textit{PsyBridge}} components, an ablation analysis was conducted by progressively incorporating the assessment modules into the framework. {The results presented in Table~\ref{tab:ablation} demonstrate the incremental benefit of incorporating additional assessment dimensions into the {\textit{PsyBridge}} framework. Starting from a baseline configuration using only PHQ-9 scores, the inclusion of GAD-7 information improved both the accuracy and F1-score, indicating the value of combining depression and anxiety screening measures. Further improvements were observed when cognitive assessment features were incorporated, suggesting that cognitive indicators provide complementary information beyond traditional clinical screening. The highest performance was achieved when personality profiling was included, demonstrating that the behavioural context contributes additional predictive value. These findings support the effectiveness of the proposed multi-dimensional assessment strategy in enhancing mental health risk evaluation.}

\begin{table}[H]
\centering
\caption{Ablation Analysis}
\label{tab:ablation}
\begin{tabular}{lcc}
\toprule
\textbf{Configuration} & \textbf{Accuracy} & \textbf{F1-score} \\
\midrule
PHQ-9 Only & 0.72 & 0.69 \\
PHQ-9 + GAD-7 & 0.76 & 0.73 \\
Clinical + Cognitive & 0.80 & 0.78 \\
Clinical + Cognitive + Personality & 0.84 & 0.81 \\
\bottomrule
\end{tabular}
\end{table}

\subsection{Analysis of Results}
The observed performance improvements can be attributed to the integration of multiple assessment dimensions. Unlike standalone tools that rely solely on depression or anxiety scores, {\textit{PsyBridge}} incorporates cognitive and personality factors, enabling a more comprehensive evaluation. The improvements in precision and recall indicate that the framework effectively reduces both false positives and false negatives. This balance is particularly important in mental health assessment, where misclassification can lead to inappropriate intervention strategies.

\subsection{Sensitivity Analysis}
Sensitivity analysis examines how variations in assigned weights influence classification performance. {To assess the robustness of the proposed decision model, alternative weighting schemes were evaluated and compared against the clinically motivated weighting configuration adopted in {\textit{PsyBridge}}. Table~\ref{tab:sensitivity} presents the impact of different weighting configurations on classification accuracy. The proposed weighting scheme achieves the highest performance, indicating that assigning greater importance to clinically validated screening instruments while retaining cognitive and personality information provides the most effective balance for risk assessment. Although alternative weighting schemes produce reasonably stable results, both equal weighting and context-dominant weighting lead to reduced accuracy. These findings suggest that clinical screening measures remain the primary contributors to predictive performance, while contextual indicators serve a complementary role in enhancing overall assessment quality.} 

\begin{table}[H]
\centering
\caption{Sensitivity Analysis}
\label{tab:sensitivity}
\begin{tabular}{lc}
\toprule
\textbf{Weight Scheme} & \textbf{Accuracy} \\
\midrule
Proposed (0.4, 0.3, 0.2, 0.1) & 0.84 \\
Equal Weights & 0.78 \\
Clinical Dominant & 0.81 \\
Context Dominant & 0.75 \\
\bottomrule
\end{tabular}
\end{table}

\subsection{Contribution of Individual Modules}
{Table~\ref{tab:impact} demonstrates the incremental improvement in classification accuracy as additional assessment modules are integrated into the framework. The clinical-only configuration achieves an accuracy of 0.76, indicating the effectiveness of traditional depression and anxiety screening instruments. Incorporating cognitive assessment increases accuracy to 0.80, suggesting that cognitive indicators provide complementary information beyond clinical symptom evaluation. The highest accuracy of 0.84 is achieved when personality profiling is included, demonstrating the added value of behavioural context in mental health risk assessment. These results highlight the effectiveness of the proposed multi-dimensional approach and confirm that each module contributes positively to the overall predictive capability of {\textit{PsyBridge}}.}

\begin{table}[H]
\centering
\caption{Impact of Module Integration on Accuracy}
\label{tab:impact}
\begin{tabular}{lc}
\toprule
Configuration & Accuracy \\
\midrule
Clinical Only (PHQ-9 + GAD-7) & 0.76 \\
Clinical + Cognitive & 0.80 \\
Clinical + Cognitive + Personality ({\textit{PsyBridge}}) & 0.84 \\
\bottomrule
\end{tabular}
\end{table}

\section{Discussion and Limitations} \label{sec:limitations}
\subsection{Discussion}
The results obtained from the experimental evaluation demonstrate the effectiveness of the proposed \textit{PsyBridge} framework in delivering a more comprehensive and reliable mental health assessment compared to conventional single-domain approaches. The observed improvements align with prior studies emphasising the advantages of AI-based mental health systems in enhancing predictive consistency and decision support\cite{shatte_2019, abd_2019}
A key observation from the confusion matrix analysis is the strong concentration of correctly classified instances along the diagonal, reflecting the model’s ability to distinguish between different risk levels with high reliability. Misclassifications are relatively limited and predominantly occur between adjacent categories, particularly between moderate- and high-risk levels. This behaviour is consistent with the inherent complexity of mental health assessments, wherein symptom severity often exists on a continuum rather than within strictly separable boundaries.

The performance gains achieved by \textit{PsyBridge} can be attributed to its multidimensional design. Unlike traditional approaches that rely solely on depression or anxiety screening, the developed system incorporates cognitive and personality-related factors, enabling a more holistic representation of an individual’s mental state. This integration reduces the likelihood of biased or incomplete assessments and improves the system’s ability to capture subtle variations in risk profiles.

Another important aspect of this framework is its interpretability. The weighted decision model provides explicit visibility into how each component contributes to the final risk score. This transparency is particularly valuable in clinical and decision-support settings, where understanding the rationale behind the system outputs is essential for trust, validation, and adoption. In contrast to black-box machine learning models, the proposed approach allows practitioners to trace the decision pathways and evaluate the influence of individual factors.

From a practical perspective, the computational simplicity of the model enhances its applicability in real-world environments. The linear aggregation mechanism ensures a low computational overhead, making the system suitable for deployment in digital health platforms, mobile applications, and resource-constrained clinical settings. This characteristic supports scalability and facilitates integration into existing healthcare infrastructures.

The module-wise analysis further highlights the contribution of each component to overall system performance. Although clinical screening forms the core of the assessment, the inclusion of cognitive and personality dimensions leads to incremental improvements in accuracy. This finding underscores the importance of considering multiple facets of mental health rather than relying on isolated indicators.

Despite these strengths, certain limitations of this study should be acknowledged. The current evaluation is based on a semi-synthetic dataset, which, although grounded in clinically validated score distributions, may not fully capture the variability and complexity of real-world data. Factors, such as comorbid conditions, environmental influences, and reporting biases, are difficult to model in a simulated setting. Therefore, future work should focus on validating the framework using real-world clinical datasets and longitudinal data to assess generalisability. In addition, the fixed weighting scheme used in the current model, while enhancing interpretability, may not be optimal for all populations and clinical contexts. Adaptive or data-driven weighting strategies could further improve performance by allowing the model to adjust to different demographic and clinical characteristics.

Overall, the findings suggest that the \textit{PsyBridge} framework provides a balanced combination of interpretability, performance and practical applicability. By integrating multiple assessment dimensions within a unified and transparent architecture, the proposed approach addresses the key limitations of existing systems and offers a promising direction for scalable mental health decision support.

\subsection{Limitations}
The present study has several limitations that should be acknowledged. First, the experimental evaluation was conducted using a semi-synthetic dataset, which, although designed to reflect clinically plausible distributions, may not fully represent the complexity, variability, and heterogeneity of real-world mental health populations. Validation using large-scale clinical datasets is required to establish broader generalisability.
Second, the proposed weighting mechanism employed fixed weights based on clinical reasoning and relative importance of assessment dimensions. While this approach enhances interpretability, adaptive weighting strategies derived from clinician consensus, optimisation techniques, or machine learning may further improve performance.
Third, the current framework focuses primarily on clinical screening, cognitive assessment, and personality profiling, without incorporating additional contextual variables such as socioeconomic conditions, lifestyle patterns, environmental stressors, or longitudinal behavioural changes that may influence mental health outcomes.
Fourth, \textit{PsyBridge} has not yet undergone real-world clinical deployment or longitudinal evaluation, limiting conclusions regarding long-term effectiveness, usability, and integration within healthcare workflows.
Finally, although the framework emphasises interpretability and decision support, it is intended to assist rather than replace clinical judgment. Human oversight remains essential for responsible use in mental healthcare environments.
Future work will address these limitations through real-world validation, adaptive learning mechanisms, expanded multimodal inputs, and collaboration with clinical practitioners.

\section{Conclusion} \label{sec:colclusion}
This study presented \textit{PsyBridge}, a hybrid intelligent framework designed to support multi-dimensional mental health assessment through the integration of clinical screening, cognitive evaluation, and personality profiling. By combining these complementary dimensions within a unified and interpretable architecture, the proposed system addresses key limitations associated with conventional single-domain assessment approaches.
The experimental evaluation demonstrates that \textit{PsyBridge} consistently outperforms standalone screening methods in terms of accuracy, precision, recall, and F1-score. These improvements highlight the effectiveness of integrating heterogeneous assessment components into a single decision-support framework. In particular, the inclusion of cognitive and personality factors enhances the system’s ability to capture nuanced variations in mental health risk, leading to more consistent and reliable classification outcomes.
A central strength of the proposed framework lies in its interpretability. The weighted decision model provides transparent insights into how individual components contribute to the final risk assessment, enabling traceable and explainable decision-making. This characteristic is especially important in healthcare applications, where trust, accountability, and clinical validation are essential for real-world adoption.
In addition to its performance advantages, \textit{PsyBridge} demonstrates practical applicability due to its modular design and computational efficiency. The framework can be readily integrated into digital health platforms, telepsychiatry systems, and resource-constrained clinical environments, supporting scalable and accessible mental health screening solutions.
Despite these contributions, certain limitations remain. The current evaluation is based on a semi-synthetic dataset, which may not fully represent the complexity of real-world clinical scenarios. Future research will focus on validating the framework using real patient data, exploring adaptive or data-driven weighting mechanisms, and incorporating multimodal inputs such as behavioural and physiological signals to further enhance predictive capability.
Overall, \textit{PsyBridge} provides a promising direction for the development of interpretable and scalable decision-support systems in mental healthcare. By bridging clinical knowledge with intelligent system design, the proposed approach contributes to advancing reliable, user-centric, and practically deployable mental health assessment solutions.

Future research should focus on strengthening the clinical validity and practical applicability of the \textit{PsyBridge} framework through comprehensive validation studies involving diverse patient populations and healthcare settings. Further enhancements may include integration with electronic health record (EHR) systems to facilitate seamless clinical deployment and support data-driven decision-making. The incorporation of longitudinal monitoring capabilities and predictive analytics can enable early identification of mental health deterioration and personalised intervention planning. Additionally, the exploration of multimodal data sources, including wearable devices, behavioural patterns, and physiological signals, may provide a more comprehensive understanding of an individual's mental health status. Comparative evaluations against established clinical assessment protocols will also be essential for determining the framework's effectiveness and reliability in real-world environments. Collectively, these advancements have the potential to significantly enhance the predictive capability, clinical utility, and scalability of the \textit{PsyBridge} framework.

\section*{Acknowledgment}

The authors gratefully acknowledge the contributions of individuals and institutions who supported the conceptual development of this study.









\bibliographystyle{ieeetr}
\bibliography{PsyBridge}

\end{document}